\documentclass[11pt,letterpaper]{article}
\newcommand{\SCFPreprint}{}
\usepackage[margin=1in]{geometry}
\usepackage[T1]{fontenc}
\usepackage{lmodern}
\usepackage[round]{natbib}

\usepackage{amsmath,amssymb,amsthm}
\usepackage{booktabs,tabularx,graphicx,microtype,enumitem}
\usepackage{caption}
\usepackage{longtable}
\usepackage{float}
\usepackage{algorithm}
\usepackage{algpseudocode}
\usepackage{xcolor,url,hyperref}
\hypersetup{colorlinks=true,allcolors=blue!50!black,pdftitle={Source-balanced capped Fisher: Bounded Evidence Aggregation for User Inference in Language Models},pdfauthor={Yang Liu, Ran Zou, Wanrong Zhu}}
\newtheorem{theorem}{Theorem}

\newcommand{\ind}{\mathbf{1}}
\newcommand{\E}{\mathbb{E}}

\newcommand{\AUC}{\operatorname{AUROC}}

\newcommand{\Qhat}{\widehat Q}
\newcommand{\Rcal}{\mathcal R}
\newcommand{\Ccal}{\mathcal C}

\newcommand{\calD}{\mathcal{D}}
\newcommand{\calE}{\mathcal{E}}
\newcolumntype{Y}{>{\raggedright\arraybackslash}X}

\title{Source-balanced Capped Fisher: Bounded Evidence Aggregation for User Inference in Language Models}
\author{
Yang Liu\textsuperscript{1}
\qquad
Ran Zou\textsuperscript{2}
\qquad
Wanrong Zhu\textsuperscript{2,\thanks{Corresponding author: wanronz1@uci.edu}}
\\[3pt]
\textsuperscript{1}Department of Statistics and Operations Research,\\
University of North Carolina at Chapel Hill
\\
\textsuperscript{2}Department of Statistics, University of California, Irvine
}
\date{}

\begin{document}
\maketitle

\begin{abstract}
User inference asks whether a source contributed documents to a language-model update, even when the auditor observes only fresh documents from that source. Document-level scores can provide evidence of participation, but combining this evidence is challenging: documents differ in baseline predictability and in how strongly they reflect information learned from other documents of the same source. We propose source-balanced capped Fisher (SCF), an aggregation method that uses known nonparticipating sources to assess how unusual each document score is. It transforms the resulting reference-tail frequencies into negative-log evidence, caps individual contributions, and averages them within each source. This limits the influence of isolated extreme scores while retaining evidence across multiple documents. We complete the inference procedure with a decision rule that tests source participation using a p-value calibrated against a separate set of known nonparticipating sources. The rule provides finite-sample Type-I error control under source exchangeability. Experiments across four real-world datasets show that SCF achieves the highest AUROC among the evaluated aggregation methods in four domains. With the calibration-based decision rule, SCF delivers consistently competitive detection power across the tested settings, achieving the highest power in many cases, while empirical false-positive rates remain generally close to or below the nominal levels.

\end{abstract}
\section{INTRODUCTION}
\label{sec:introduction}

Suppose a news organization, Alder Press, wants to know whether its articles were used in a language-model update (Figure~\ref{fig:source-motivation}). It cannot inspect the training corpus, but it can evaluate the updated model's predictions on articles from its own collection. Even if none of these auditing articles were used in the update, the model's predictions may reflect topics or writing patterns learned from other articles by the same organization. 
This is an instance of \emph{user inference} \citep{kandpal2024user}. A user, which we refer to as a \emph{source}, is a collection of documents associated with a common origin, such as an author, a social-media account, a developer, or a mailbox owner. Given a language model after an update and query documents from a target source, the task is to determine whether at least one document from that source was included in the update’s training data. We focus on the setting in which the query documents themselves were not used in the update, so the inference concerns participation through other documents from the same source.

\ifdefined\SCFPreprint
\begin{figure}[H]\else\begin{figure}[ht]\fi
  \centering
\ifdefined\SCFPreprint\includegraphics[width=0.48\linewidth]{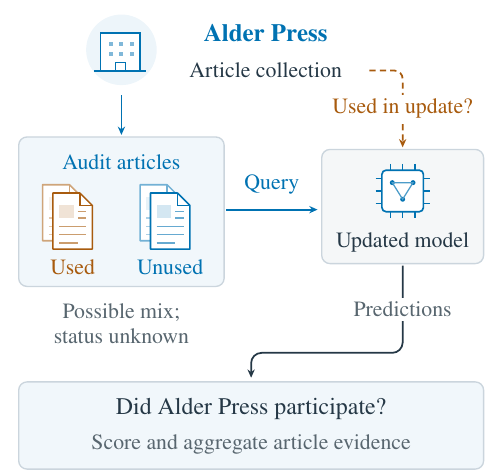}\else\includegraphics[width=0.8\columnwidth]{figures/source_inference_motivation.pdf}\fi
  \caption{Example of user inference. Alder Press asks whether any of its articles entered a model update. It queries the updated model with its own articles, which may all be unused in fine-tuning. A fixed scoring rule and source-level aggregation yield participation evidence.}
  \label{fig:source-motivation}
\end{figure}

A more extensively studied question is whether an individual example was used for training, known as membership inference \citep{shokri2017membership}. Existing approaches include likelihood-ratio attacks such as LiRA \citep{carlini2022membership} and methods that calibrate membership scores for example difficulty \citep{watson2022difficulty}.  For language models, document-level scores have been constructed using comparisons with neighboring texts \citep{mattern2023neighborhood}, selected or normalized token statistics \citep{shi2024detecting,zhang2025minkpp}, and calibration against corpus token frequencies \citep{zhang2024dcpdd}. These approaches provide document-level scores, characterizing model behavior on individual documents. However, nonmembership of a queried document does not imply nonparticipation of its source. Dataset inference \citep{maini2024dataset,zhao2025posthoc} and collection-level membership inference \citep{puerto2025scaling} extend this analysis by combining evidence across multiple inspected documents. Although these approaches aggregate document evidence, their target remains the training use of the suspect collection. Our question differs: we seek to infer participation through other documents from the same source, even when every inspected document is absent from the update.

Existing approaches to user inference include RaMIA \citep{tao2025range}, which aggregates record-level membership scores using trimmed averages, and the language-model method of \citet{kandpal2024user}, which averages document-level log-likelihood differences between fine-tuned and reference models.   Our focus is on effectively aggregating document-level scores into an informative source-level statistic and calibrating it for a participation decision. We use reference data from sources known not to have participated in the update to guide this aggregation and calibration. Such reference sources may be available, for example, through platform accounts created after the update, when participation is defined at the account level. The reference sources provide a baseline for interpreting score values under nonparticipation. Our procedure accommodates different document-level scoring rules, including likelihood differences and scores computed from the target model alone. Whether a reference model is required therefore depends on the chosen scoring rule.

We propose \emph{source-balanced capped Fisher} (SCF). Given a document-level scoring rule, SCF constructs a reference distribution of scores that balances contributions across known nonparticipating sources. The same scoring rule is applied to target and reference documents, with scores oriented so that larger values favor participation. For each query document, SCF computes the upper-tail frequency of its score under this reference distribution. Smaller frequencies indicate scores that are more unusual relative to the reference sources. SCF then takes the negative logarithms of these frequencies, caps the resulting contributions, and averages them across the target source's query documents. The resulting source-level statistic combines evidence on a reference scale while bounding each document's contribution.

We complete the inference procedure with a decision rule that turns the aggregated score into a participation decision. Controlling the
probability of falsely declaring participation is essential and requires a valid null comparison \citep{zhang2025cannot}. We obtain a rank-based p-value \citep{bates2023testing} by comparing
the target's aggregated score with those of a separate set of known nonparticipating calibration sources. We reject the nonparticipation hypothesis when the p-value is at most the chosen significance level. Under source-level exchangeability, this rule provides finite-sample Type-I error control.

Our contributions are twofold. First, we propose SCF, a new source-level score that aggregates
document evidence through a reference-tail transformation and capped log averaging. Across four real-world domains, SCF achieves the highest mean AUROC among the evaluated aggregation methods using the same document-level scores.
Second, we complete source participation inference with a decision rule based on  rank calibration. The rule applies to any fixed source-level scoring rule and provides finite-sample Type-I error control under source-level exchangeability. Across the evaluated methods, empirical false-positive rates
remain generally close to or below the nominal levels. SCF delivers consistently competitive detection power across the tested settings, achieving the highest mean power in many cases.

\subsection{Additional related work}
\label{sec:additional-related-work}

\paragraph{User inference beyond modern LLMs.}
Early work studied user-level data auditing in text-generation models \citep{song2019auditing}. User-level inference has also been investigated in automatic speech recognition \citep{miao2021audio} and facial recognition \citep{chen2023face}, where the queried recordings or images need not themselves have been used for training. Subject membership inference studies a related question in federated learning, including settings in which a subject's records are distributed across multiple clients \citep{suri2022subject}. More broadly, distributional membership inference concerns participation by a data-generating party without requiring knowledge of its exact training records \citep{hartmann2023distribution}. These formulations share our interest in participation by a common source, although their model-access assumptions and inference procedures differ.

\paragraph{Calibration for membership and user inference.}
Statistical auditing also requires translating evidence into a decision with an appropriate interpretation of errors. MembershipTracker empirically calibrates user-level mean losses against nonmember users in a proactive data-marking setting \citep{chen2025membershiptracker}. For record-level membership inference, MIAFdR uses conformal calibration and multiple testing to control false discoveries under its calibration assumptions \citep{zhao2025miafdr}. More generally, \citet{bates2023testing} study reference-based conformal $p$-values for outlier detection, including their dependence and use in multiple testing. Randomized data-use auditing provides another basis for statistical decisions: \citet{huang2024datause} combine prior randomized marking with sequential testing to bound false detections. Our focus is the calibration of a source-level statistic constructed from multiple query documents, using reference data from sources known not to have participated in the update.

\section{USER INFERENCE FROM FRESH QUERIES}
\label{sec:problem}

\paragraph{Sources and hypothesis.}

A source is a collection of documents associated with a common origin, such as a user or a website. For each source $u$, let $\calE_u$ denote its document collection. We assume that these collections are disjoint, i.e., $\calE_u\cap\calE_v=\varnothing$ for $u\ne v$. Consider a language-model update from an initial model $B$ to an updated model $M$, using a training document set $\calD_0$. User inference asks whether any document from source $u$ was used in this update:

\begin{equation}
H_{0,u}:|\calD_0\cap\calE_u|=0,
\qquad
H_{1,u}:|\calD_0\cap\calE_u|>0.
\label{eq:hypothesis}
\end{equation}
These hypotheses concern participation in the audited update, regardless of whether the source contributed to earlier training of $B$.

\paragraph{Query documents.}

To infer participation of $u$, the auditor observes a collection of query documents $\mathbf X_u=(x_{u1},\ldots,x_{um_u})$, where $m_u\ge1$ and $x_{uj}\in\calE_u$. We focus on the fresh-query setting, in which none of the query documents belongs to $\calD_0$. Even in this setting, training on other documents from the same source may change the model's predictions on the observed queries, providing evidence of source participation.

\paragraph{Document-level scores.}
For any document $x$, let $S(x)$ be a document score, with larger values providing stronger evidence of corresponding source participation. Our aggregation method accepts any such prespecified score. Following the reference-model log-likelihood comparison used for user inference by \citet{kandpal2024user}, we use a token-normalized log-likelihood difference:

\begin{equation}
 S(x)=\frac{\log\widetilde p_{M}(x)-\log\widetilde p_{B}(x)}{n_x},
 \label{eq:ld}
\end{equation}
where $n_x$ is the number of scored next-token prediction positions in the document $x$, and $\widetilde p_B(x)$ and $\widetilde p_M(x)$ are the products of the corresponding token probabilities under $B$ and $M$ over those positions. The score measures the per-token improvement in predictive fit after the update. Comparing with $B$ accounts for baseline predictability, while dividing by $n_x$ puts documents on a per-token scale. \citet{kandpal2024user} do not use normalization, but we found that using \eqref{eq:ld} can enhance the performance of every method we test. To transform these document scores into source-level evidence and calibrate the resulting statistic, we
use sources known not to have participated in the
update.

\paragraph{Known nonparticipating sources.}
We assume access to sources known not to have contributed any documents to the update. We divide these nonparticipating sources into two disjoint groups, using $r\in\Rcal$ to index a reference source and $c\in\Ccal$ to index a calibration source, with $K=|\Rcal|$ and $L=|\Ccal|$. Each source $v\in\Rcal\cup\Ccal$ satisfies $\calD_0\cap\calE_v=\varnothing$, and supplies a query collection $\mathbf X_v=(x_{v1},\ldots,x_{vm_v})$, where $m_v\ge1$ and $x_{vj}\in\calE_v$. We use $\Rcal$ to construct the document-evidence transformation and $\Ccal$ to calibrate the resulting source-level statistic, as described in Section~\ref{sec:method}.

\section{SOURCE-BALANCED CAPPED FISHER  SCORE}
\label{sec:method}

Starting from the document-level scores, we construct a measure of source participation by comparing each query document with documents from known nonparticipating sources and combining the resulting evidence within a source. Figure~\ref{fig:overview} illustrates the procedure. We first use the reference sources to assess how unusual each document score is relative to nonparticipating sources. We then limit the contribution of individual documents and average their evidence to obtain a source-level statistic. A separate calibration step, described later, converts this statistic into a participation decision.

\ifdefined\SCFPreprint\begin{figure}[H]\else\begin{figure*}[t]\fi
\centering
\includegraphics[width=0.8\textwidth]{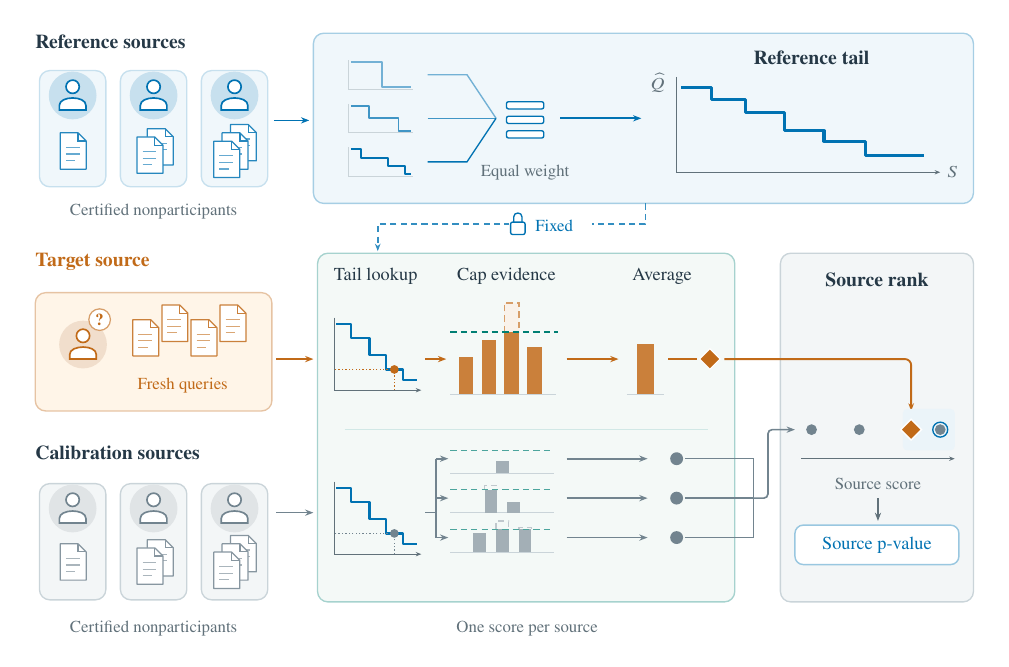}
\caption{SCF bounds document evidence before source-level calibration. Certified nonparticipating reference sources contribute equally to a fitted upper-tail map. The frozen map is applied to fresh queries from the target and each calibration source; log-tail evidence is capped (dashed lines) and averaged within each source. The target score (orange diamond) is ranked against calibration scores (gray circles) to obtain a source-level p-value; the shaded tail marks calibration scores at least as large as the target score. All icons and curves are schematic.}
\label{fig:overview}
\ifdefined\SCFPreprint\end{figure}\else\end{figure*}\fi

\subsection{Constructing the source-level score}
\label{sec:source-score}

\paragraph{Reference scale.}

A raw document score measures improvement of predictive fit after
the update, but interpreting its magnitude requires a reference scale. We construct this scale using the query documents of the nonparticipating reference sources $\Rcal$. For a threshold $z$, define

\begin{equation}
 \overline Q_K(z)=\frac1K\sum_{r\in\Rcal}\frac1{m_r}\sum_{j=1}^{m_r}\ind\{S(x_{rj})\ge z\}.
\end{equation}

Within each reference source, we calculate the proportion of query documents whose scores are at least $z$, then average these proportions across sources.

To avoid zero tail frequencies, we apply a small smoothing adjustment:

\begin{equation}
 \Qhat(z)=\frac{1+N\overline Q_K(z)}{N+1},
 \label{eq:qhat}
\end{equation}
where $N=\sum_{r\in\Rcal}m_r$. This keeps $\Qhat(z)$ positive even when $z$ exceeds every reference score. 

For a query document $x$, a small value of $\Qhat(S(x))$ indicates that its score is unusually large relative to the nonparticipating reference sources.

\paragraph{Combining document evidence.}
We further express the unusualness of a query document as $-\log\Qhat(S(x))$, so that smaller reference-tail frequencies produce larger evidence values. A natural source-level summary is the average of these values across its query documents. However, a document with a very small reference-tail frequency can contribute disproportionately to this average. To limit its influence while retaining evidence from multiple documents, we cap each contribution at $-\log\alpha$, for $\alpha\in(0,1)$. Then, for source $u$ with query collection $\mathbf X_u$, we define the source-balanced capped Fisher (SCF) score as
\begin{equation}
 SC_\alpha(\mathbf X_u)=-\frac1{m_u}\sum_{j=1}^{m_u}\log\max\{\Qhat(S(x_{uj})),\alpha\}.
 \label{eq:method}
\end{equation}
We also include $\alpha=0$ as the uncapped case, with the convention $-\log0=+\infty$. Larger values of $SC_\alpha(\mathbf X_u)$ indicate stronger evidence of source participation.  
 
For $\alpha>0$, documents with reference-tail frequencies below $\alpha$ receive the same maximum contribution. The effect of capping is that multiple moderately unusual documents can outweigh a single extreme document, as increasing the score of an already extreme document no longer increases its contribution.  For illustration, consider two sources with reference-tail frequencies $(.15,.15)$ and $(.01,1)$. Their uncapped scores are approximately $1.897$ and $2.303$, respectively. With $\alpha=.05$, the first score remains unchanged, while the second falls to $1.498$. Capping therefore allows two moderately unusual documents to outweigh one extreme document.

The negative-log aggregation in \eqref{eq:method} follows the form of Fisher's combination rule. Related methods such as TFisher select and weight small-$p$-value contributions \citep{zhang2018tfisher}, whereas SCF caps the largest log-tail contributions. Bounding individual contributions follows a classical principle of robust estimation \citep{huber1964robust}; DC-PDD uses a related cap on calibrated token contributions before averaging them into a document score \citep{zhang2024dcpdd}. We use empirical reference-tail frequencies in this construction and calibrate the resulting statistic at the source level.

\paragraph{Computation.}
Equation~\eqref{eq:method} requires $\widehat Q$ only at the observed scores $S(x_{u1}),\ldots,S(x_{um_u})$. For each query document, substitute $z=S(x_{uj})$ into the reference-tail calculation and then apply the smoothing in~\eqref{eq:qhat}. Averaging the resulting capped log-tail contributions gives $SC_\alpha(\mathbf X_u)$. Thus we need only finitely many evaluations, rather than an explicit representation of $\widehat Q$ at every real threshold. Direct calculation for Algorithm~\ref{alg:scf} costs $O(m_uN)$ comparisons with no additional language-model queries.

\begin{algorithm}[H]
\caption{Source-balanced capped Fisher (SCF)}
\label{alg:scf}
\begin{algorithmic}[1]
\Require Cached scores $\{S(x_{ri})\}_{i=1}^{m_r}$ for $r\in \Rcal$,  and $\{S(x_{uj})\}_{j=1}^{m_u}$; cap $\alpha\in[0,1)$.
\Ensure Source score $SC_\alpha(\mathbf X_u)$.
\State $N\gets\sum_{r\in\Rcal}m_{r}$
\For{$j=1,\ldots,m_u$}
    \State $z\gets S(x_{uj})$
    \State $\displaystyle \overline Q_K(z)\gets\frac1K\sum_{r\in\Rcal}\frac1{m_{r}}\sum_{i=1}^{m_{r}}\mathbf1\{S(x_{ri})\ge z\}$
    \State $\widehat Q(z)\gets\{1+N\overline Q_K(z)\}/(N+1)$
\EndFor
\State \Return $\displaystyle -\frac1{m_u}\sum_{j=1}^{m_u}\log\max\{\Qhat(S(x_{uj})),\alpha\}$
\end{algorithmic}
\end{algorithm}

\subsection{Testing source participation}
\label{sec:participation-test}

The source-level statistic $SC_\alpha(\mathbf X_u)$ summarizes evidence of participation, with larger values indicating stronger evidence. To decide whether a target source $u$ participated in the update, we compare its statistic with those of the known nonparticipating calibration sources $\Ccal$, which are separate from the reference sources $\Rcal$ used to construct the score. The rank-based p-value is defined as
\begin{equation}
 p_u=\frac{1+\sum_{c\in\Ccal}\ind\{SC_\alpha(\mathbf X_c)
                       \ge SC_\alpha(\mathbf X_u)\}}{L+1}.
 \label{eq:conformal}
\end{equation}

A small $p_u$ indicates that few calibration sources have scores at least as large as the target's. At a prespecified significance level $t$, we reject $H_{0,u}$ when $p_u\le t$. The significance level $t$ controls the rejection threshold and is distinct from the evidence-cap parameter $\alpha$.

The following theorem establishes that this rank-based quantity is a valid p-value under source-level exchangeability.

\begin{theorem}[Finite-sample source-level validity]
\label{thm:source-validity}
Under $H_{0,u}$, suppose the source statistics $SC_\alpha(\mathbf X_u)$ and $SC_\alpha(\mathbf X_c)$, $c\in\Ccal$, are jointly exchangeable. Then, for every $t\in[0,1]$,
\begin{equation}
 \Pr_{H_{0,u}}(p_u\le t )
 \le  t,
 \label{eq:validity-bound}
\end{equation}
where $p_u$ is defined in~\eqref{eq:conformal}.
\end{theorem}

\section{EXPERIMENTS}
\label{sec:experiments}

We evaluate the two stages of user inference: assessing the quality of our SCF score and calibrating it for a decision. The experiments evaluate Reddit \citep{baumgartner2020pushshift}, CC News \citep{Hamborg2017}, Enron \citep{klimt2004enron}, and Chromium \citep{meyers2018dataset, chang2020convokit} under a common audit protocol. 

\subsection{Experimental setup}
\label{sec:experimental-setup}

\paragraph{Data and model update.} We use four datasets with different source and document units: Reddit accounts and their comments, CC News web domains and their articles, Enron mailbox owners and their sent emails, and Chromium developer accounts and their comments. The first three corpora also appear in User Inference paper \citep{kandpal2024user}, while Chromium extends the comparison to developer communications. For each dataset, the initial model $B$ is pretrained GPT-Neo-125M \citep{black_sid_2021_5297715, gao2020pile}. 
We obtain the updated model $M$ by fine-tuning $B$ on documents from designated participating sources. 
Appendix~\ref{app:experiments} provides full details on fine-tuning and selection of training data (participants).

\paragraph{Methods compared.}
We compare SCF with four existing aggregated scores, which we refer to as Mean, Max, Min \citep{kandpal2024user}, and RaMIA \citep{tao2025range}. These scores compute the mean, maximum, minimum, and average of the highest-75\% of document scores, respectively, using the token-normalized likelihood difference defined in \eqref{eq:ld}.
For SCF, we fix the cap at $\alpha=.05$ across all four domains, and we additionally include \emph{Uncapped Fisher}, which is the special case $\alpha=0$ in SCF. The ablation study varies only $\alpha$ to see the effect of cap strength.

\paragraph{Source partitions.}
For each audit partition, we divide the nonparticipating sources into reference sources \(\mathcal R\), calibration sources \(\mathcal C\), and nonparticipating test sources. We use \((K,L)=(40,59)\) for Reddit, CC News, and Chromium, and \((25,25)\) for Enron since Enron has limited sources. Note that the four baseline methods require no reference fitting and therefore use all sources in  \(\mathcal R\cup\mathcal C\) for calibration. All participating sources and the nonparticipating test sources form the test set. In the test set, each source contributes at most 16 query documents, all excluded from fine-tuning. Thus, the audit tests whether a source participated using documents distinct from those used in the model update.  

We repeat the audit over 500 partitions of the nonparticipating sources, keeping the model weights, training participation, query documents, and document scores unchanged.

\paragraph{Evaluation metrics.}
We use AUROC to compare the source-level scores produced by different aggregation methods and assess which method best distinguishes participating from nonparticipating sources. For a scalar source score $G$, AUROC is the probability that a randomly selected participating source receives a higher score than an independently selected nonparticipating source, with half credit for ties \citep{fawcett2006introduction}. Let $\mathbf X_1$ and $\mathbf X_0$ be independent participant and nonparticipant query collections, independent of the reference. Conditional on the reference, 
\begin{equation}
\begin{aligned}
 \AUC(G\mid\Rcal)
 &=\Pr\{G(\mathbf X_1)>G(\mathbf X_0)\mid\Rcal\}\\
 &\quad+\tfrac12\Pr\{G(\mathbf X_1)=G(\mathbf X_0)\mid\Rcal\}.
\end{aligned}
\label{eq:auc}
\end{equation}
We evaluate participation decisions using TPR and FPR at various significance levels, and further compare the different methods.
At a chosen significance level $t$, the true-positive rate (TPR) is the proportion of participating test sources for which $p_u\le t$, and the false-positive rate (FPR) is the corresponding proportion among nonparticipating test sources. AUROC evaluates discrimination across score thresholds, while TPR and FPR evaluate the calibrated decision rule at a specified level.
The code used to reproduce our aggregation and calibration results are available online.\footnote{\url{https://github.com/yangliu6768/scf-reproduction}}

\begin{table*}[htbp]
\centering
\small
\setlength{\tabcolsep}{5pt}
\caption{Source-level AUROC under a common control budget, with fixed GPT-Neo-125M models, document scores, and query bags. Entries average 500 source-role partitions with identical test sources across methods. SCF uses $(K,L)=(40,59)$ for Reddit, CC News, and Chromium, and $(25,25)$ for Enron, with $\alpha=.05$. AUROC is computed before calibration. Bold denotes the highest mean AUROC.}
\label{tab:domains}
\begin{tabular}{@{}lrrrrrrr@{}}
\toprule
Domain & Mean & Max & Min & RaMIA & Uncapped Fisher & SCF \\
\midrule
Reddit & 0.6443 & 0.6176 & 0.6158 & 0.6444 & 0.6438 & \textbf{0.6456}\\
CC News & 0.6305 & 0.5953 & 0.6050 & 0.6227 & 0.6327 & \textbf{0.6354}\\
Enron & 0.8068 & 0.7222 & 0.6104 & 0.7973 & 0.8227 & \textbf{0.8309}\\
Chromium & 0.6178 & 0.5527 & 0.5620 & 0.6196 & 0.6279 & \textbf{0.6346}\\
\bottomrule
\end{tabular}
\end{table*}

\subsection{Comparing aggregation methods via AUROC}

Table~\ref{tab:domains} shows that SCF attains the highest mean AUROC among the evaluated aggregators in all four domains. The means summarize the 500 source-role partitions for the fixed models and source pools. The AUROC of SCF is .6456 on Reddit, .6354 on CC News, .8309 on Enron, and .6346 on Chromium. Mean is the strongest raw-score baseline among Mean, Max, Min, and RaMIA on CC News and Enron, while RaMIA is strongest on Reddit and Chromium. Relative to them, the gains are .12, .50, 2.41, and 1.50 percentage points, respectively. SCF also exceeds uncapped Fisher at the common cap $\alpha=.05$ in every domain. 

The baseline ordering also helps interpret these comparisons. Mean and RaMIA both outperform Max and Min in every domain, indicating that combining multiple document scores is more effective here than using either extreme alone. However, the ordering of Mean and RaMIA changes across domains.

Table~\ref{tab:paired-counts-effects} in the appendix reports the complete paired win, tie, and loss counts and mean AUROC gains against each comparator. Against the strongest raw-score baseline by AUROC, SCF wins in all 500 partitions on CC News, in $98.4\%$ on Enron, and in $87.8\%$ on Chromium. On Reddit, its smaller $.12$-point mean gain is accompanied by a $61.4\%$ strict win rate.

\subsection{Effect of the evidence cap}
\label{sec:cap-ablation}

To isolate the effect of capping, we keep the fitted tail map, queries, and source partitions fixed and compare uncapped Fisher with $\alpha\in\{.01,.02,.05,.10\}$. Figure~\ref{fig:cap-ablation} examines sensitivity to the cap. Every positive cap in the tested grid has higher mean AUROC than the uncapped condition in each domain. The dependence on cap strength nevertheless differs across domains.  Increasing $\alpha$ from $.05$ to $.10$ further improves CC News, Enron, and Chromium; on Enron, mean AUROC increases from $.8309$ to $.8356$. Reddit instead peaks at $\alpha=.02$ with AUROC $.6460$, declining to $.6450$ at $\alpha=.10$. Thus, increasing the cap parameter need not improve source ordering even though it further limits every document's maximum contribution. These results support the use of bounded contributions in the evaluated settings, while leaving data-driven cap selection open. Appendix~\ref{app:cap-ablation} reports absolute AUROCs and variation across partitions.

Table~\ref{tab:cap-mechanism} clarifies what evidence is affected. At $\alpha=.05$, the source-averaged cap-activation rate is approximately $5\%$ among nonparticipants, compared with $6.40\%$--$9.43\%$ among participants. Extreme-tail occurrences are therefore more frequent among participants in these experiments. SCF preserves this distinction: each affected document retains the contribution $-\log\alpha$, so multiple such documents can still increase a source's average evidence. Capping limits the additional weight assigned to tail depth while retaining the contribution of an extreme event. The matched AUROC improvements in Table~\ref{tab:domains} support this bounded treatment without implying that extreme-tail occurrences themselves lack participation information.

\begin{table}[htbp]
\centering
\caption{Frequency of capped document evidence at $\alpha=.05$. Entries are percentages of query documents satisfying $\widehat Q(S(x))<.05$, averaged within each evaluated source, and then over the same 500 source-role partitions as Table~\ref{tab:domains}. Reference and calibration sources are excluded.}
\label{tab:cap-mechanism}
\begin{tabular}{@{}lrr@{}}
\toprule
Domain & Nonparticipant (\%) & Participant (\%)\\
\midrule
Reddit & 5.180 & 8.270\\
CC News & 5.188 & 9.430\\
Enron & 4.943 & 9.398\\
Chromium & 5.244 & 6.401\\
\bottomrule
\end{tabular}
\end{table}

\ifdefined\SCFPreprint\begin{figure}[H]\else\begin{figure}[t]\fi
\centering
\ifdefined\SCFPreprint\includegraphics[width=0.65\linewidth]{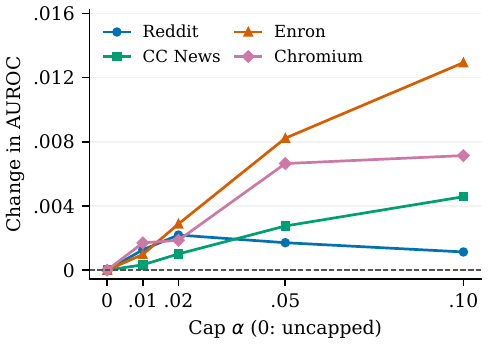}\else\includegraphics[width=\columnwidth]{figures/alpha_ablation.pdf}\fi
\caption{Mean AUROC gain over uncapped Fisher, averaged over 500 matched source-role partitions. Here $\alpha=0$ retains reference smoothing and omits the explicit cap.}
\label{fig:cap-ablation}
\end{figure}

\subsection{Source-level decisions and false-positive rates}
\label{sec:fpr}

We compute a conservative source-level conformal rank as in~\eqref{eq:conformal} and declare participation when $p_u\le t$, for $t\in\{.05,.10,.20\}$. Each rank uses the calibration panel specified in Section~\ref{sec:experimental-setup}; raw-score baselines use the union of the reference and SCF calibration controls.  Table~\ref{tab:fpr} reports TPR and FPR over the same 500 partitions as Table~\ref{tab:domains}, with the SCF cap fixed at $\alpha=.05$.

\begin{table*}[!t]
\centering
\caption{Source-level audit utility at nominal level $t$. TPR and FPR (\%) are averaged over the same 500 source-role partitions as Table~\ref{tab:domains}. SCF uses $\alpha=.05$. Bold marks the highest mean TPR for each domain and nominal level.}
\label{tab:audit-utility}
\label{tab:fpr}
\small
\setlength{\tabcolsep}{5pt}
\renewcommand{\arraystretch}{1.06}
\begin{tabular*}{\ifdefined\SCFPreprint\linewidth\else0.6\textwidth\fi}{@{\extracolsep{\fill}}lrlrrrrr@{}}
\toprule
Domain & $t$ (\%) & Metric & Mean & Max & Min & RaMIA & SCF \\
\midrule
Reddit & 5 & TPR & 9.53 & 3.08 & 4.59 & \textbf{10.73} & 8.82 \\
 &  & FPR & 4.95 & 4.86 & 4.95 & 5.07 & 4.95 \\
\addlinespace[3pt]
 & 10 & TPR & \textbf{17.70} & 12.04 & 16.08 & 17.46 & 17.21 \\
 &  & FPR & 10.08 & 10.04 & 9.84 & 9.99 & 10.02 \\
\addlinespace[3pt]
 & 20 & TPR & \textbf{33.60} & 24.87 & 32.32 & 31.16 & 32.62 \\
 &  & FPR & 19.95 & 19.99 & 19.70 & 19.90 & 20.05 \\
\midrule
CC News & 5 & TPR & 13.17 & 8.01 & 11.47 & 12.20 & \textbf{14.14} \\
 &  & FPR & 5.13 & 5.04 & 5.16 & 5.14 & 4.86 \\
\addlinespace[3pt]
 & 10 & TPR & 20.87 & 13.83 & 19.12 & 19.55 & \textbf{24.08} \\
 &  & FPR & 10.23 & 10.03 & 10.08 & 10.25 & 10.08 \\
\addlinespace[3pt]
 & 20 & TPR & 34.33 & 28.87 & 32.73 & 33.24 & \textbf{36.62} \\
 &  & FPR & 20.16 & 19.96 & 20.10 & 20.19 & 20.29 \\
\midrule
Enron & 5 & TPR & 4.66 & \textbf{15.09} & 1.55 & 4.63 & 11.32 \\
 &  & FPR & 4.08 & 3.89 & 3.60 & 4.08 & 3.96 \\
\addlinespace[3pt]
 & 10 & TPR & \textbf{41.60} & 31.06 & 19.51 & 30.05 & 32.92 \\
 &  & FPR & 10.06 & 9.71 & 9.60 & 10.18 & 7.81 \\
\addlinespace[3pt]
 & 20 & TPR & 68.77 & 48.50 & 44.72 & 68.63 & \textbf{70.09} \\
 &  & FPR & 19.37 & 19.66 & 19.42 & 19.91 & 19.10 \\
\midrule
Chromium & 5 & TPR & 5.97 & 4.14 & 6.32 & 4.60 & \textbf{8.87} \\
 &  & FPR & 5.22 & 4.93 & 5.08 & 5.11 & 5.40 \\
\addlinespace[3pt]
 & 10 & TPR & 17.86 & 6.74 & 10.62 & 19.98 & \textbf{21.83} \\
 &  & FPR & 10.48 & 9.74 & 10.30 & 10.42 & 10.68 \\
\addlinespace[3pt]
 & 20 & TPR & 34.46 & 25.21 & 25.40 & 34.58 & \textbf{36.50} \\
 &  & FPR & 20.34 & 20.03 & 20.12 & 20.39 & 20.30 \\
\bottomrule
\end{tabular*}
\end{table*}

SCF has the highest mean TPR at all three levels on CC News and Chromium, and at the 20\% level on Enron. At $t=.10$ on CC News, SCF increases TPR from $20.87\%$ for Mean to $24.08\%$, a gain of $3.21$ percentage points, while its FPR is slightly lower ($10.08\%$ versus $10.23\%$). On Chromium at the same nominal level, TPR increases from $19.98\%$ for RaMIA to $21.83\%$, accompanied by an FPR increase from $10.42\%$ to $10.68\%$. At $t=.20$, however, SCF has both higher TPR ($36.50\%$ versus $34.58\%$) and slightly lower FPR ($20.30\%$ versus $20.39\%$) than RaMIA. The ranking gains therefore translate into improved audit utility in many settings, although SCF does not attain the highest TPR in every domain and at every level.

AUROC and calibrated detection consequently provide complementary assessments. Enron has the largest AUROC gain over the strongest raw-score baseline, yet SCF does not have the highest TPR at every evaluated level. Reddit's smaller AUROC gain likewise does not yield the highest TPR at these levels. The ranking results show better average pairwise ordering across the evaluated partitions, while Table~\ref{tab:fpr} shows where that improvement translates into greater detection at a specified nominal level. Therefore, both comparisons are needed to assess audit utility. An additional experiment trains only on documents preceding a
temporal cutoff and uses later documents for all target, reference, and calibration queries; SCF retains detection power with empirical FPRs close to or below the nominal levels (Appendix~\ref{app:temporal-audit}).

\section{CONCLUSION}
\label{sec:conclusion}

We studied user inference from query documents that were excluded from the audited model update. SCF addresses the aggregation problem by mapping fixed document scores to a source-balanced reference scale and averaging capped log-tail evidence. This construction limits the influence of individual documents while retaining evidence that recurs across a source. A separate source-level calibration step provides finite-sample Type-I error control under source exchangeability, allowing arbitrary dependence among documents within each source.

Across four domains with fixed GPT-Neo-125M models and document scores, SCF attains the highest mean AUROC among the evaluated aggregators. Matched comparisons with uncapped Fisher show that capping improves mean ranking at a common cap. The higher cap-activation rates among participants support retaining extreme-tail occurrences as evidence while limiting the additional weight assigned to their depth. The decision results complement these ranking comparisons by showing where the gains translate into higher detection rates at specified nominal levels. Together, these findings support bounded aggregation as a useful approach to combining document evidence for user inference. Source-level calibration then turns the resulting score into a participation decision with an explicit statistical guarantee under the stated assumptions.
\section{FUTURE WORK}
\label{sec:future-work}

Selecting the best cap from available audit controls remains an open problem. Our experiments use a common $\alpha=.05$ and examine sensitivity over a fixed grid; they do not establish an optimal choice for population AUROC. A theoretical direction is to characterize when participation signals admit effective cap selection using only nonparticipating controls. These controls inform the null evidence distribution, whereas the AUROC effect of capping also depends on how participation changes the frequency and depth of tail events. Developing a selection principle therefore calls for explicit assumptions connecting these quantities. Under such assumptions, future work could seek bounds on the ranking loss of a selected cap relative to the best fixed cap in a specified class.

A second direction concerns audits under source-population shift. Distribution shifts between member and nonmember examples, including temporal shifts, can confound evaluations of LLM membership inference~\citep{duan2024membership}. The current validity result assumes exchangeability of the null target and calibration sources. Extending the audit to mismatched populations requires an explicit model of that mismatch and a quantitative account of its effect on false positives. Building on conformal inference beyond exchangeability~\citep{barber2023conformal}, one could study sensitivity bounds that accommodate dependence within each source and uncertainty about the discrepancy between source populations. The aim would be an audit whose reported error guarantee reflects the allowed degree of mismatch, making the population assumptions underlying its conclusion explicit.

Also, the empirical scope can be broadened through more backbones and domains. The current comparisons use one document-score rule and one fitted GPT-Neo-125M model \citep{black_sid_2021_5297715, gao2020pile} per domain. Crossing aggregation rules with other document scores, such as DC-PDD \citep{zhang2024dcpdd} and Min-K\%++ \citep{zhang2025minkpp}, model sizes, training runs, query budgets, and other domains would further support the generalization.

\label{maintextend}
\clearpage
\section*{AI Use Statement}
Generative AI tools assisted with cleaning and reformatting datasets, refining language, and writing proofs. We have reviewed all AI-assisted work and take responsibility for the final content of this work.
\bibliography{references}
\clearpage
\appendix
\section{IMPLEMENTATION CONVENTIONS}
\label{app:algorithm}

All query bags are nonempty and all document scores are finite. Reference identifiers are distinct and disjoint from calibration and target identifiers. If $z$ is no larger than every reference score, $\Qhat(z)=1$; above all reference scores, $\Qhat(z)=1/(N+1)$. Thus reference smoothing makes the uncapped score finite. For $\alpha>0$, each document contribution is additionally bounded by $C=-\log\alpha$. Computation uses double precision, the resulting tails lie in $[1/(N+1),1]$, and a stable sum divided by the actual bag size. The ablation also evaluates the smoothed uncapped score $SC_0$.

\section{PROOF OF SOURCE-LEVEL CONFORMAL VALIDITY}
\label{app:calibration}

\begin{proof}[Proof of Theorem~\ref{thm:source-validity}]
Work under $H_{0,u}$. Write
$Z_0=SC_\alpha(\mathbf X_u)$ and $Z_\ell=SC_\alpha(\mathbf X_{C_\ell})$ for $\ell=1,\ldots,L$. 

For each $i=0,\ldots,L$, define the conservative descending rank
\[
 \rho_i=\sum_{j=0}^L\ind\{Z_j\ge Z_i\}.
\]
Equation~\eqref{eq:conformal} gives $p_u=\rho_0/(L+1)$. For any realized score vector and integer $k\in\{0,\ldots,L+1\}$,
\[
 \sum_{i=0}^L\ind\{\rho_i\le k\}\le k.
\]
Indeed, ordering the scores from largest to smallest assigns every member of a tied group the rank of its last position. Only the first $k$ positions can have rank at most $k$. This covers complete ties and $k=0$.

Exchangeability gives the $L+1$ indicators the same expectation. With $k=\lfloor t(L+1)\rfloor$,
\begin{align*}
 \Pr_{H_{0,u}}(p_u\le t)
 =\frac1{L+1}\E_{H_{0,u}}\!\left[
   \sum_{i=0}^L\ind\{\rho_i\le k\}\right]
 \le\frac{k}{L+1}\le t.
\end{align*}
\end{proof}

\section{EXPERIMENTAL DETAILS}
\label{app:experiments}

\subsection{Source populations and query construction}

Table~\ref{tab:inventory} summarizes the eligible audit-source pool after query filtering; training exposure in Table~\ref{tab:training} covers the full fine-tuning corpus. Source participation is fixed by the training assignment: participating sources contribute fine-tuning documents, while certified nonparticipating sources contribute none. Query documents are disjoint from the fine-tuning and checkpoint-validation documents.

\begin{table}[H]
\centering
\small
\setlength{\tabcolsep}{4pt}
\caption{Source populations and query bags. The SCF reference/calibration allocation is $(K_d,L_d)=(40,59)$ in Reddit, CC News, and Chromium and $(25,25)$ in Enron. All participants and all null sources outside these controls are tested in each of 500 partitions. Raw-score baselines use all $K_d+L_d$ controls for calibration. Query counts are per source.}
\label{tab:inventory}
\begin{tabular}{@{}llrrrrr@{}}
\toprule
Domain & Source unit & Sources & Participants & Null pool & Null test & Queries \\
\midrule
Reddit & Account & 267 & 84 & 183 & 84 & 16 \\
CC News & Web domain & 2,709 & 1,348 & 1,361 & 1,262 & 1--16 \\
Enron & Mailbox owner & 141 & 70 & 71 & 21 & 1--16 \\
Chromium & Developer & 199 & 50 & 149 & 50 & 16 \\
\bottomrule
\end{tabular}
\end{table}

The query budget is $m_u=\min\{16,q_u\}$, where $q_u$ is the number of eligible query documents for source $u$. The same query bag is used in every partition and by every aggregation method. CC News has 2,087 sources with fewer than 16 queries and an overall median bag size of six; Enron has 51 sources with fewer than 16 queries and an overall median bag size of 16. Reddit and Chromium each have 16 queries for every source. The shared budget limits the number of documents rather than the total number of tokens.

\begin{itemize}[leftmargin=*,itemsep=3pt]
\item \textbf{Reddit.} Sources are public accounts, with text drawn from two communities per account. Each participating account contributes eight fine-tuning documents per community. Query bags contain eight documents from each community. Views retain 8--511 content tokens and append EOS, giving the same range of prediction-token counts.
\item \textbf{CC News.} Sources are web domains and documents are articles. Each scored view uses the first at most 511 content tokens with an appended EOS. An article is excluded if any fixed-width or paragraph segment within its first 2,044 content tokens exactly matches a complete token input in the training-candidate or validation pool, including unused training candidates. Segment inputs include EOS; comparisons use token-sequence hashes. This is an exact-input exclusion, not a near-duplicate criterion. The accompanying \path{config/data_construction.md} specifies segment boundaries and short-tail handling. All aggregation methods score the same first-block article view.
\item \textbf{Enron.} Sources are mailbox owners identified by a canonical sender address. Within each mailbox, candidate From addresses contain the surname inferred from the mailbox name; the most frequent candidate is selected, with lexical tie-breaking. Only messages sent by that address within its owner mailbox are retained, and repeated canonical addresses define one source. Before source assignment, line-ending-normalized message bodies are deduplicated within sources; any exact body shared by multiple sources is removed from all of them. Sources with fewer than four remaining documents are excluded. Each email view retains at most 511 content tokens and appends EOS. Exact token matches with the update or validation inputs are excluded before applying the query budget; this removes two overlapping query documents and leaves 8,104 eligible queries across the 141 sources.
\item \textbf{Chromium.} Sources are developer accounts and documents are comments. Quote-prefixed lines are removed, and eligible views contain 65--257 content tokens, giving 64--256 next-token predictions. No EOS is appended. Each participating account contributes 16 training documents; each source's 16 queries are drawn by the common rule from its 32 eligible query documents.
\end{itemize}

\subsection{Model fitting and checkpoint selection}

All domains use \texttt{EleutherAI/gpt-neo-125m} as both the initial model and the pretrained scoring reference. Fine-tuning uses Adam with learning rate $5\times10^{-5}$, batch size eight, $(\beta_1,\beta_2)=(.9,.999)$, $\epsilon_{\rm Adam}=10^{-8}$, and zero weight decay. The schedule has 200 warmup updates followed by linear decay over the ten-epoch training horizon. Documents are not packed, and no gradient accumulation or clipping is used. Training uses bfloat16 autocast with float32 optimizer state. The training seeds are \texttt{20260914} for Reddit and Chromium,
\texttt{171470688381003029} for CC News, and
\texttt{7129716279750356844} for Enron. Each seed initializes the Python and PyTorch CPU/CUDA random states and a separate PyTorch generator for shuffling training documents; NumPy uses the seed
modulo $2^{32}$. These seeds are distinct from the audit source-role partitions, whose exact assignments are included in the public repository.

We choose the epoch endpoint with minimum validation token-average negative log-likelihood, breaking exact ties in favor of the earlier endpoint. Validation uses held-out documents from participating sources. The numbers of validation documents are 336, 17,191, 2,636, and 200 for Reddit, CC News, Enron, and Chromium, respectively. Table~\ref{tab:training} reports the resulting checkpoints and training exposure.

\begin{table}[H]
\centering
\small
\caption{Training exposure and checkpoint used for scoring. The checkpoint criterion is minimum validation token-average negative log-likelihood over ten epoch endpoints.}
\label{tab:training}
\begin{tabular}{lrrrr}
\toprule
Domain & Train documents / epoch & Prediction tokens / epoch & Epoch & Updates \\
\midrule
Reddit & 1,344 & 62,965 & 3 & 504 \\
CC News & 283,370 & 95,590,331 & 9 & 318,798 \\
Enron & 42,917 & 10,742,290 & 5 & 26,825 \\
Chromium & 800 & 84,258 & 2 & 200 \\
\bottomrule
\end{tabular}
\end{table}

\subsection{Document scores and aggregation}

For a document view with predicted tokens $w_t$ and fixed contexts, the evidence is
\[
 S(x)=\frac1{n_x}\sum_{t\in\mathcal T_x}
 \{\log p_M(w_t\mid w_{<t})-\log p_B(w_t\mid w_{<t})\},
 \qquad n_x=|\mathcal T_x|.
\]
Here $\mathcal T_x$ is the set of scored prediction positions, and $p_M,p_B$ correspond to the same updated and initial checkpoints as~\eqref{eq:ld}; $\widetilde p_M(x)$ and $\widetilde p_B(x)$ are the products over those positions. The target and reference models use identical tokens and prediction masks. Larger scores indicate stronger participation evidence, and all aggregators receive the same numeric document scores. In domain $d$, SCF fits one source-weighted tail map from $K_d$ reference bags in each partition and uses it for both calibration and test bags. We use $(K_d,L_d)=(40,59)$ for Reddit, CC News, and Chromium and $(25,25)$ for Enron, where $L_d$ is SCF's calibration count. Every reference source has total weight $1/K_d$, including when its bag size differs from other sources. With equal bag sizes, this gives the same tail as pooled document counts. The main comparisons use $\alpha=.05$; Appendix~\ref{app:cap-ablation} varies the cap while retaining the same reference map within each partition.

Mean, Max, and Min aggregate the numerical scores $S(x_{uj})$ directly. For the RaMIA-style statistic, order the query documents so that $S(x_{u(1)})\ge\cdots\ge S(x_{u(m_u)})$ and define
\[
 G_{\rm trim}(\mathbf X_u)=\frac1{k_{m_u}}\sum_{j=1}^{k_{m_u}}S(x_{u(j)}),
 \qquad k_{m_u}=\max\{1,\lfloor.75m_u\rfloor\}.
\]
We use the .25 trimming ratio from the CelebA identity-range experiment of \citet{tao2025range}. This evaluates RaMIA's aggregation component on the shared document score. Uncapped Fisher uses the same source-balanced, smoothed reference map as SCF, with $\alpha=0$. The four raw-score baselines require no fitted reference map and use all $K_d+L_d$ controls for calibration; SCF and uncapped Fisher reserve $K_d$ controls for reference fitting and use the remaining $L_d$ for calibration.

\subsection{Source partitions and AUROC}

For each domain $d$ and a random partition $b$, we form the fitting reference $\mathcal R_{db}$ and the SCF calibration panel $\mathcal C_{db}$. All remaining null sources and all participating sources form the common test panel. Reference, SCF calibration, and test roles are disjoint within each partition. For the raw-score baselines, the calibration set is the union $\mathcal C^{\rm all}_{db}=\mathcal R_{db}\cup\mathcal C_{db}$. Thus every method receives the same total budget of $K_d+L_d$ controls and the same test targets. The participating/null test counts are 84/84, 1,348/1,262, 70/21, and 50/50 for Reddit, CC News, Enron, and Chromium, respectively.

We use 500 partition seeds. Training membership, model weights, query documents, and document scores remain fixed across partitions; only the null-source audit roles change.

Let $\mathcal B_{\mathrm{raw}}=\{\mathrm{Mean},\mathrm{Max},\mathrm{Min},\mathrm{RaMIA}\}$ denote the four raw-score comparators. For method $a$ in domain $d$, let $\widehat A^a_{db}$ be its source-score AUROC in partition $b$, with half credit for equal scores. Table~\ref{tab:domains} reports
\[
\overline A^a_d=\frac{1}{500}\sum_{b=1}^{500}\widehat A^a_{db}.
\]

Table~\ref{tab:paired-counts-effects} complements the mean AUROC comparison with paired win, tie, and loss counts against each comparator. Within each partition, all methods are evaluated on the same test sources, with SCF fixed
at $\alpha=.05$. 

\begin{table}[htbp]
\centering
\caption{Paired AUROC comparisons across 500 matched source-role partitions.
Wins, ties, and losses count partitions in which SCF ($\alpha=.05$) has higher,
equal, or lower AUROC than the named comparator, respectively; each row sums
to 500. Mean gain is the average paired AUROC difference (SCF minus comparator),
expressed in percentage points and computed before rounding.}
\label{tab:paired-counts-effects}
\small
\setlength{\tabcolsep}{7pt}
\renewcommand{\arraystretch}{1.08}
\begin{tabular}{@{}llrrrr@{}}
\toprule
Domain & Comparator & Wins & Ties & Losses & Mean gain (pp) \\
\midrule
Reddit & Mean & 329 & 10 & 161 & $+0.130329$ \\
 & Max & 460 & 0 & 40 & $+2.798073$ \\
 & Min & 450 & 0 & 50 & $+2.976502$ \\
 & RaMIA & 307 & 4 & 189 & $+0.117630$ \\
 & Uncapped Fisher & 336 & 5 & 159 & $+0.170607$ \\
\midrule
CC News & Mean & 500 & 0 & 0 & $+0.496548$ \\
 & Max & 500 & 0 & 0 & $+4.007021$ \\
 & Min & 500 & 0 & 0 & $+3.041496$ \\
 & RaMIA & 500 & 0 & 0 & $+1.271364$ \\
 & Uncapped Fisher & 498 & 0 & 2 & $+0.274898$ \\
\midrule
Enron & Mean & 492 & 1 & 7 & $+2.409660$ \\
 & Max & 499 & 0 & 1 & $+10.874694$ \\
 & Min & 500 & 0 & 0 & $+22.055374$ \\
 & RaMIA & 489 & 2 & 9 & $+3.358367$ \\
 & Uncapped Fisher & 390 & 15 & 95 & $+0.822177$ \\
\midrule
Chromium & Mean & 473 & 3 & 24 & $+1.679760$ \\
 & Max & 500 & 0 & 0 & $+8.189760$ \\
 & Min & 491 & 0 & 9 & $+7.257600$ \\
 & RaMIA & 439 & 1 & 60 & $+1.496080$ \\
 & Uncapped Fisher & 404 & 10 & 86 & $+0.664000$ \\
\bottomrule
\end{tabular}
\end{table}

\subsection{Source-level true- and false-positive rates}
\label{app:fpr}

Table~\ref{tab:fpr} uses the same models, queries, reference fits, source-role partitions, and test panels as Table~\ref{tab:domains}. Let $G_{db}^{(a)}$ be method $a$'s source score in domain $d$ and partition $b$: $SC_{.05,\mathcal R_{db}}$ for SCF, $SC_{0,\mathcal R_{db}}$ for uncapped Fisher, and the appropriate aggregation of $S(\mathbf X_u)$ for a raw-score baseline. Define the method-specific calibration set and its size by
\[
 \mathcal C_{db}^{(a)}=
 \begin{cases}
  \mathcal C_{db}^{\rm all}, & a\in\mathcal B_{\rm raw},\\
  \mathcal C_{db}, & a\in\{\mathrm{SCF},\mathrm{Uncapped\ Fisher}\},
 \end{cases}
 \qquad L_d^{(a)}=|\mathcal C_{db}^{(a)}|.
\]
Motivated by the use of held-out calibration in conformal inference
\citep{vovk2005algorithmic,shafer2008tutorial,bates2023testing,zou2026conformalized}, we compare each test source's score with calibration scores computed using the same frozen rule, giving
\[
 p_{u,db}^{(a)}=
 \frac{1+\sum_{c\in\mathcal C_{db}^{(a)}}
       \ind\{G_{db}^{(a)}(\mathbf X_c)\ge G_{db}^{(a)}(\mathbf X_u)\}}
      {L_d^{(a)}+1},
 \qquad \text{reject if }p_{u,db}^{(a)}\le t.
\]
Comparisons share the nominal level and total control budget. Under the appropriate source exchangeability, Appendix~\ref{app:calibration} gives the validity argument for each method. 

Let $\mathcal U_{y,db}$ be the test sources of participation status $y\in\{0,1\}$. We compute the empirical class-specific rejection rate and its partition average as
\[
 \widehat R^a_{y,db}(t)=
 \frac1{|\mathcal U_{y,db}|}\sum_{u\in\mathcal U_{y,db}}
       \ind\{p_{u,db}^{(a)}\le t\},
 \qquad
 \overline R^a_{y,d}(t)=\frac1{500}\sum_{b=1}^{500}
       \widehat R^a_{y,db}(t).
\]
The $y=0$ rate is FPR and the $y=1$ rate is TPR. Controls are excluded from both denominators: calibration uses known nonparticipating sources, whereas FPR is measured on disjoint null test sources, which is common in conformal outlier detection and model auditing
\citep{bates2023testing,zhao2025miafdr,liu2026explanations}. Repeated decisions remain dependent and are not treated as independent binomial trials. Table~\ref{tab:fpr} reports empirical TPR/FPR means. 

\subsection{Evidence-cap ablation}
\label{app:cap-ablation}

The ablation uses the same 500 source-role partitions, allocations, models, document scores, and query bags as the main comparisons. Thus $(K_d,L_d)=(40,59)$ in Reddit, CC News, and Chromium, and $(25,25)$ in Enron. Within each partition, one source-balanced reference map, one calibration panel, and one test panel are shared by $\alpha\in\{0,.01,.02,.05,.10\}$. The uncapped condition is
\[
SC_0(\mathbf X_u)=\frac{1}{m_u}\sum_{j=1}^{m_u}-\log\widehat Q\{S(x_{uj})\}.
\]
The finite-reference smoothing is retained in every condition, so $\widehat Q\ge1/(N+1)>0$ even without a cap. Larger positive $\alpha$ gives a smaller maximum contribution $-\log\alpha$ and hence stronger capping. The $\alpha=0$ and $\alpha=.05$ predictions reproduce their main-table counterparts. 

Let $\widehat A_{db}(\alpha)$ be the source AUROC in domain $d$ and partition $b$. Table~\ref{tab:alpha_summary} reports its mean and sample standard deviation across partitions:
\[
\overline A_d(\alpha)=\frac{1}{500}\sum_{b=1}^{500}\widehat A_{db}(\alpha),
\qquad
s_d(\alpha)=\left\{\frac{1}{499}\sum_{b=1}^{500}\bigl[\widehat A_{db}(\alpha)-\overline A_d(\alpha)\bigr]^2\right\}^{1/2}.
\]

\begin{table}[H]
\centering
\caption{Cap sensitivity over the same 500 source-role partitions. All entries are mean AUROC $\pm$ partition SD. These SDs describe variation in source-role assignment, not population confidence intervals.}
\label{tab:alpha_summary}
\small
\setlength{\tabcolsep}{5pt}
\begin{tabular}{lccccc}
\toprule
Domain & $\alpha=0$ (uncapped) & $\alpha=.01$ & $\alpha=.02$ & $\alpha=.05$ & $\alpha=.10$ \\
\midrule
Reddit & $0.6438 \pm 0.0248$ & $0.6451 \pm 0.0248$ & $0.6460 \pm 0.0248$ & $0.6456 \pm 0.0247$ & $0.6450 \pm 0.0258$ \\
CC News & $0.6327 \pm 0.0022$ & $0.6330 \pm 0.0020$ & $0.6337 \pm 0.0018$ & $0.6354 \pm 0.0017$ & $0.6372 \pm 0.0017$ \\
Enron & $0.8227 \pm 0.0462$ & $0.8237 \pm 0.0453$ & $0.8256 \pm 0.0444$ & $0.8309 \pm 0.0428$ & $0.8356 \pm 0.0419$ \\
Chromium & $0.6279 \pm 0.0320$ & $0.6296 \pm 0.0318$ & $0.6298 \pm 0.0314$ & $0.6346 \pm 0.0310$ & $0.6351 \pm 0.0314$ \\
\bottomrule
\end{tabular}
\end{table}

All columns use the same AUROC scale. Figure~\ref{fig:cap-ablation} displays the matched change
\[
\overline D_d(\alpha)=\frac{1}{500}\sum_{b=1}^{500}\bigl[\widehat A_{db}(\alpha)-\widehat A_{db}(0)\bigr].
\]
These standard deviations describe variation in audit roles for the fixed fitted model and source pool. The partitions share sources, so they are not independent training replications and the standard deviations are not population confidence intervals.

All four positive caps have higher mean AUROC than uncapped Fisher in each domain. The highest mean in the tested grid occurs at $.02$ for Reddit and $.10$ for the other three domains.

For every cap, it is evaluated at nominal levels of 5\%, 10\%, and 20\%. Table~\ref{tab:cap-fpr} reports the resulting empirical FPR means.

\begin{table}[ht]
\centering
\small
\setlength{\tabcolsep}{4pt}
\caption{Empirical FPR (\%), averaged over the same 500 partitions. Each cap uses the same fitted reference and calibration sources, with $(K,L)=(40,59)$ except Enron $(25,25)$. The nominal level $t$ and evidence cap $\alpha$ are distinct parameters.}
\label{tab:cap-fpr}
\begin{tabular}{@{}lrrrrrr@{}}
\toprule
Domain & $t$ (\%) & Uncapped & $\alpha=.01$ & $\alpha=.02$ & $\alpha=.05$ & $\alpha=.10$ \\
\midrule
Reddit & 5 & 5.07 & 5.06 & 5.05 & 4.95 & 5.00 \\
Reddit & 10 & 10.09 & 10.11 & 10.04 & 10.02 & 9.99 \\
Reddit & 20 & 20.05 & 20.06 & 20.04 & 20.05 & 19.94 \\
CC News & 5 & 4.91 & 4.88 & 4.86 & 4.86 & 4.98 \\
CC News & 10 & 10.00 & 10.00 & 10.06 & 10.08 & 10.21 \\
CC News & 20 & 20.32 & 20.38 & 20.37 & 20.29 & 20.25 \\
Enron & 5 & 3.90 & 3.92 & 3.99 & 3.96 & 3.98 \\
Enron & 10 & 8.05 & 7.93 & 7.93 & 7.81 & 7.90 \\
Enron & 20 & 18.78 & 18.79 & 18.92 & 19.10 & 18.99 \\
Chromium & 5 & 5.22 & 5.25 & 5.32 & 5.40 & 5.26 \\
Chromium & 10 & 10.33 & 10.50 & 10.60 & 10.68 & 10.55 \\
Chromium & 20 & 20.14 & 20.18 & 20.21 & 20.30 & 20.52 \\
\bottomrule
\end{tabular}
\end{table}

\subsection{Auditing with post-cutoff documents}
\label{app:temporal-audit}

We evaluate SCF when the documents available to the auditor come from a later period than those used in the model update. For each domain, we do fine-tuning using only documents preceding a common temporal cutoff. All query documents from participating targets, nonparticipating targets, reference sources $\mathcal R$, and calibration sources $\mathcal C$ are dated at or after that cutoff and excluded from both fine-tuning and checkpoint validation. This setting adds temporal separation to fresh-document inference: the participation signal must transfer from earlier training documents to later queries. Calibration and null targets follow the same post-cutoff query rule; the validity guarantee continues to require the source exchangeability stated in Theorem~\ref{thm:source-validity}.

The cutoffs are December 1, 2015 for Reddit, January 1, 2018 for CC News, October 1, 2001 for Enron, and January 1, 2014 for Chromium, with timestamps normalized to UTC. Documents with missing or unparseable dates are excluded. We fit one GPT-Neo-125M model per domain with training seed \texttt{20260914}, learning rate $5\times10^{-5}$, and batch size eight, selecting the checkpoint with minimum pre-cutoff validation token-average negative log-likelihood over ten epoch endpoints. The fine-tuning sets contain 633, 148,629, 32,535, and 391 documents per epoch, respectively; the selected epochs are 3, 9, 5, and 3.

We retain $\alpha=.05$ and $(K,L)=(40,59)$, except for Enron, where $(K,L)=(25,25)$. We use the score in~\eqref{eq:ld} and the decision rule in~\eqref{eq:conformal}, averaging FPR and TPR across partitions as in Appendix~\ref{app:fpr}.

\begin{table}[t]
\centering
\small
\caption{SCF audit utility with pre-cutoff training and exclusively post-cutoff target, reference, and calibration queries. Each entry is FPR / TPR (\%), averaged over 500 source-role partitions with $\alpha=.05$. The columns specify the nominal significance level $t$.}
\label{tab:temporal-scf-utility}
\begin{tabular}{lccc}
\toprule
Domain & $t=.05$ & $t=.10$ & $t=.20$ \\
\midrule
Reddit   & 4.84 / 10.33 & 9.72 / 17.70 & 19.35 / 27.27 \\
CC News  & 5.06 / 10.43 & 10.16 / 19.88 & 20.31 / 35.55 \\
Enron    & 3.60 / 7.10 & 6.96 / 22.81 & 18.28 / 56.23 \\
Chromium & 4.90 / 8.39 & 9.93 / 15.89 & 20.18 / 28.15 \\
\bottomrule
\end{tabular}
\end{table}

Table~\ref{tab:temporal-scf-utility} shows that the calibrated SCF procedure retains detection ability under temporal separation without marked empirical false-positive inflation.

\subsection{Computational checks and reproducibility}
\label{app:reproducibility}

Fine-tuning and scoring use an NVIDIA A100 80GB GPU; cached aggregation runs on an ARM64 macOS CPU with no further model queries.

The public repository contains the data-processing and training implementations, fixed document splits, and source-role partitions for the main four-domain experiment. The complete pipeline reconstructs model inputs from the original public corpora, fine-tunes the language models, computes document scores, and evaluates the aggregation and calibration procedures. A second route starts from the supplied document-score arrays and reproduces Tables~\ref{tab:domains}--\ref{tab:fpr} and~\ref{tab:paired-counts-effects}--\ref{tab:cap-fpr} and Figure~\ref{fig:cap-ablation} without further model queries. The repository README provides the commands, software versions, and computational requirements, and a verification script checks the main tables against the values reported here.

\subsection{Responsible use}
\label{app:responsible-use}

Participation inference can expose source involvement through untrained documents. Our experiments concern controlled updates, not misconduct by named people, and report no raw personal identifiers. 

\end{document}